\documentclass{article}
\usepackage{ijcai19}

\usepackage{times}
\usepackage{soul}
\usepackage{url}
\usepackage[hidelinks]{hyperref}
\usepackage[utf8]{inputenc}
\usepackage[small]{caption}
\usepackage{graphicx}
\usepackage{amsmath}
\usepackage{booktabs}
\usepackage{algorithm}
\usepackage{algorithmic}
\usepackage{xcolor}
\usepackage{bm}
\usepackage{amsfonts}
\usepackage{mathrsfs}
\usepackage{multirow}
\usepackage{graphicx}
\usepackage{subfigure}
\usepackage{comment}

\title{STG2Seq: Spatial-temporal Graph to Sequence Model \\
for Multi-step Passenger Demand Forecasting}

\author{
Lei Bai$^1$\footnote{Contact Author}\and
Lina Yao$^1$\and
Salil.S Kanhere$^1$\And
Xianzhi Wang$^2$\and
Quan.Z Sheng$^3$
\affiliations
$^1$University of New South Wales\\
$^2$University of Technology Sydney\\
$^3$Macquarie University\\
\emails
baisanshi@gmail.com, \{lina.yao, salil.kanhere\}@unsw.edu.au,
xianzhi.wang@uts.edu.au, michael.sheng@mq.edu.au
}

\begin{document}

\maketitle

\begin{abstract}
Multi-step passenger demand forecasting is a crucial task in on-demand vehicle sharing services. 
However, predicting passenger demand over multiple time horizons is generally challenging due to the nonlinear and dynamic spatial-temporal dependencies.
In this work, we propose to model multi-step citywide passenger demand prediction based on a graph and use a hierarchical graph convolutional structure to capture both spatial and temporal correlations simultaneously. Our model consists of three parts: 1) a long-term encoder to encode historical passenger demands; 2) a short-term encoder to derive the next-step prediction for generating multi-step prediction; 3) an attention-based output module to model the dynamic temporal and channel-wise information. Experiments on three real-world datasets show that our model consistently outperforms many baseline methods and state-of-the-art models. 
\end{abstract}

\section{Introduction}
The widespread adoption of on-demand vehicle (cars and bikes) sharing services has revolutionized urban transportation. Passengers can now easily discover available vehicles in their surroundings using apps running on smartphones such as Uber, Didi, and Ola. However, these new ride-sharing platforms still suffer from the age-old problem of low accuracy in predicting passenger demand. On the one hand, drivers often have to drive a long way before they can find passengers due to low demand volumes in their proximity; on the other hand, passengers may experience long delays in obtaining rides due to high demands around their locations. This mismatch often leads to the excessive waiting time of passengers and a loss of income and waster energy resources for the drivers \cite{baipakdd}.
In particular, accurate prediction of passenger demand over multiple time steps (i.e., multi-step prediction) in different regions of the city is crucial for effective vehicle dispatching to overcome the aforementioned mismatch problem. 

\begin{figure}[htbp]
\centering
\subfigure[]{
\includegraphics[width=4cm,height=3cm]{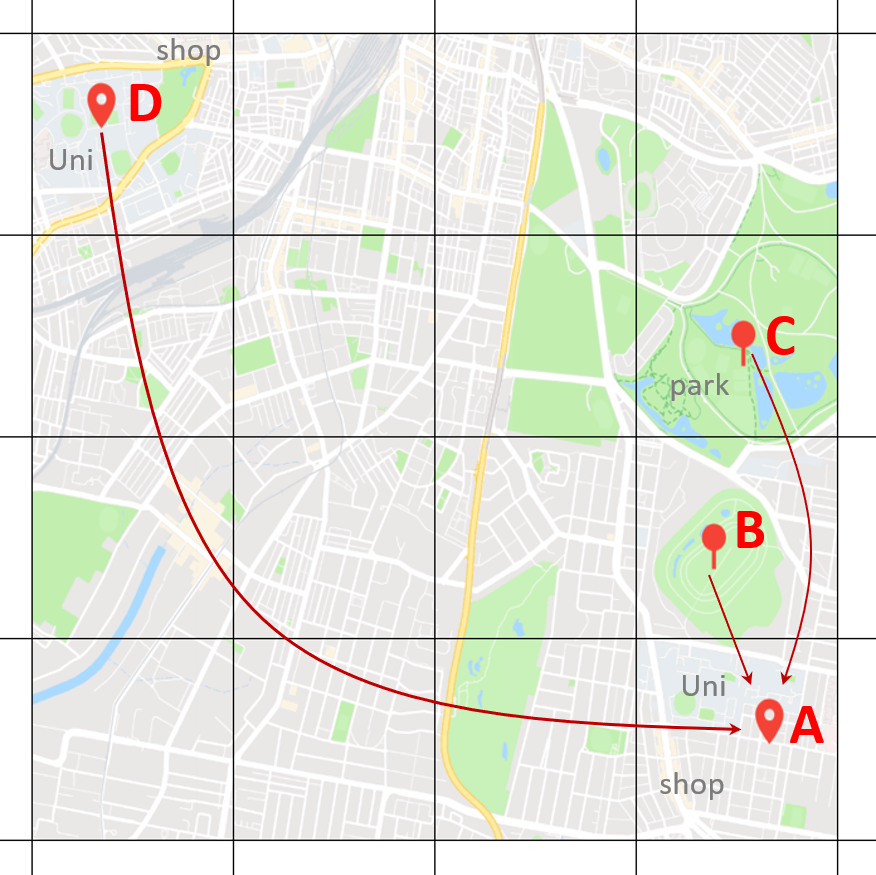}
}
\subfigure[]{
\includegraphics[width=4cm,height=3cm]{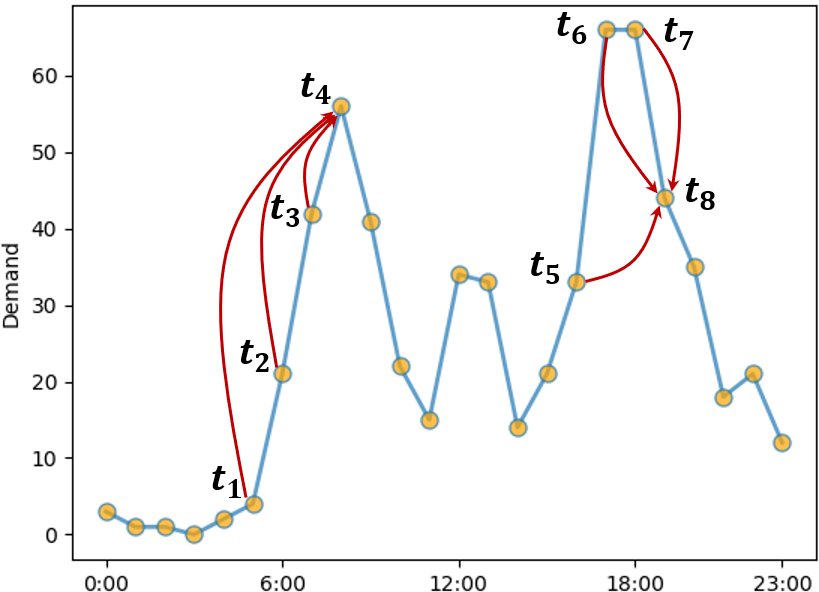}
}
\vspace{-0.2cm}
\caption{(a): Example of non-Euclidean spatial correlation; (b): Illustration of dynamic temporal correlation.}
\label{stcorrelation}
\end{figure}

Predicting passenger demand is a challenging task due to the complex, nonlinear and dynamic spatial-temporal dependencies: the future passenger demand of a target region is influenced by not only the historical demand of this region but also the demand of other regions in the city \cite{ResST-Net,DMVST}. Previous work have proposed to use time series models \cite{first}, cluster models \cite{second}, or hybrid methods \cite{thrid} to capture the correlations.
Recent works focus on leveraging the representation capabilities of deep learning methods. These methods usually employ RNN and its variants such as Long-Short Term Memory (LSTM) \cite{DMVST} networks to capture temporal correlation and CNN \cite{CNN} to extract spatial relationships from the whole city \cite{DeepST,ResST-Net} or geographically nearest regions \cite{DMVST}. Similarly, hybrid models that combine CNN and RNN (e.g., Convolutional LSTM (ConvLSTM) \cite{ConvLSTM,Jintao:short-term,AttenConvLSTM}) are proposed to extract both spatial and temporal correlations simultaneously.  

However, these methods suffer from the following drawbacks:

\begin{itemize}
    \item CNN-based methods (including ConvLSTM) assume that a city is partitioned into small grids (such as 1km $\times$ 1km area), which does not always hold \cite{FlowFlexDP}. Moreover, these methods can only model the Euclidean relationships between near regions and remote regions but not the non-Euclidean correlation among remote regions with similar characteristics. Consider the example in Figure 1(a). Region A shares points of interests with region D (university and shopping areas) rather than co-located regions B and C (which contain parks). Thus, the passenger demand in region A shows a stronger correlation with D rather than B and C.  
    \item Current methods strictly rely on RNN-based architecture (e.g., hybrid CNN-LSTM and ConvLSTM architectures) to capture temporal correlations. However, typical chain structured RNN architectures require execution of a number of iterative steps (equal to the window size of the input data) to process the demand data, and therefore
    lead to severe information oblivion in modeling the long-term temporal dependency. Moreover, utilizing RNN as a decoder for multi-step prediction is known to cause error accumulation in every step and may result in faster model deterioration \cite{STGCN}.
    \item The current research efforts do not capture the dynamics that may exist in temporal correlation accurately. 
    Most of them only reflect the collective influence of historical passenger demands.
    However, each previous step may have different and time-varying influence on the target step. 
    Figure \ref{stcorrelation} (b) illustrates a passenger demand time series for a particular region where the influence of $t_1$, $t_2$ and $t_3$ on $t_4$ varies significantly. Moreover, the importance of $t_1$, $t_2$, $t_3$ on $t_4$ is different from the importance of $t_5$, $t_6$, $t_7$ on $t_8$.
\end{itemize}

In this paper, we propose a sequence-to-sequence model for multi-step passenger demand forecasting that is based on Graph Convolutional Networks (GCN) to solve the issues above. 
Specifically, we formulate the passenger demand on a graph with each region in the city acting as a node. 
Multiple GCN layers are utilized to form a Gated Graph Convolutional Module (GGCM) to capture the spatial and temporal relationships at the same time.  
Based on the GGCM, two encoder modules named long-term encoder and short-term encoder are designed to encode historical passenger demand and integrate new predictions, separately. Compared to the chain structured RNN, the hierarchical GCN structure shortens the path to capture the long-range temporal dependency \cite{c2s}. 
Besides, having two distinct encoders allows our model to utilize last step's prediction to generate the next step's prediction without requiring a RNN to act as a decoder, which reduces the associated issue of error accumulation.
Finally, we also take into account the dynamic temporal correlations and design an attention-based output module, which can adaptively capture these dynamics. 
Overall, the contribution of this paper can be summarized as follows:
\begin{itemize}
    \item We formulate the citywide passenger demand on a graph and present a GCN-based sequence-to-sequence model for citywide multi-step passenger demand forecasting.
    To the best of our knowledge, this is the first work that purely relies on graph convolution structure to extract spatial-temporal correlations for multi-step prediction. 
    \item We propose an attention-based output module to capture the effect of the most influential historical time steps on the predicted demand and the dynamism that is inherent in these relationships.
    \item We conduct extensive experiments on three real-world datasets and compare our method with three baselines and eight discriminative deep learning based state-of-the-art methods. The experimental results show that our model can consistently outperform all the comparison methods by a significant margin.
\end{itemize}

\section{Notations and Problem Statement}
Suppose a city is partitioned into $N$ small regions, irrespective of whether grid \cite{ResST-Net} or road network \cite{FlowFlexDP} based partitioning is employed. We represent the region sets as \{$r_1$, $r_2$, ..., $r_i$, ...$r_N$\}. At each time step $t$, a 2-D matrix $\bm{D_t} \in \mathbb{R}^{N \times d_{in}}$ 
represents the passenger demand of all regions in time step $t$.
Another vector $\bm{E_t} \in \mathbb{R}^{d_e}$ represent the time features in time step $t$, which includes time of day, day of week, and information about holidays.

Given the citywide historical passenger demand sequence \{$\bm{D_0}, \bm{D_1}, ..., \bm{D_t}$\} and time features \{$\bm{E_0}, \bm{E_1}, ..., \bm{E_{t+\tau}}$\}, the target is to learn a prediction function $\bm{\Gamma}(\cdot)$ that forecasts the citywide passenger demand sequence in the next $\tau$ time steps. Instead of using all the historical passenger demand, we only consider the most recent $h$ time steps demand sequence \{$\bm{D_{t-h+1}}, \bm{D_{t-h+2}}, ..., \bm{D_t}$\} as input, which is a common practice in time series data analysis. Our problem is thus formulated as:
\begin{equation}
\begin{aligned}
    (& \bm{D_{t+1}}, \bm{D_{t+2}}, ..., \bm{D_{T}}, ..., \bm{D_{t+\tau}}) = \\
    & \bm{\Gamma}(\bm{D_{t-h+1}}, \bm{D_{t-h+2}}, ..., \bm{D_t}; \bm{E_{0}}, \bm{E_{1}}, ..., \bm{E_{t+\tau}})
\end{aligned}
\end{equation}

\section{Methodology}
\begin{figure}
\includegraphics[height=1.7in, width=3.3in]{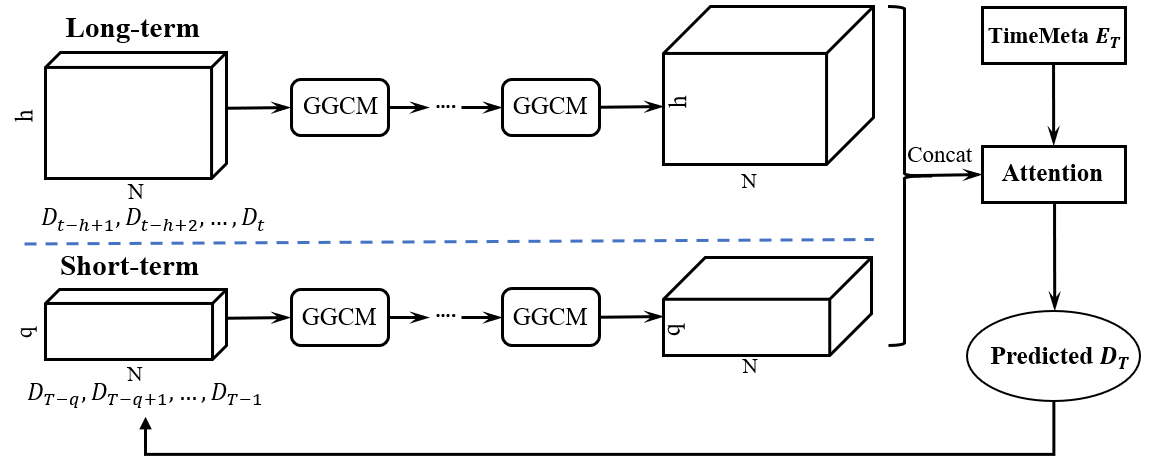}
\caption{The architecture of STG2Seq.}
\label{architecture}
\vspace{-0.2cm}
\end{figure}

The architecture of STG2Seq (Figure \ref{architecture}) comprises three components: (i) the long-term encoder (ii) the short-term encoder and (iii) the attention-based output module. Both the long-term encoder and short-term encoder comprise several serial spatial-temporal Gated Graph Convolutional Modules (GGCM), which can extract the spatial-temporal correlations simultaneously through the use of GCN along the temporal axis. We will elaborate on each component as follows. 

\subsection{Passenger Demand on Graph}
We first introduce how to formulate the citywide passenger demand on a graph. Previous works assume that the passenger demand in a region is influenced by the demand in nearby regions. However, we argue that spatial correlation does not exclusively rely on geographic locations. Remote regions may also share similar passenger demand patterns if they have similar attributes, such as point of interests (POIs).
Therefore, we treat the city as a graph $G = (\nu, \xi, A)$, where $\nu$ is the set of regions $\nu = \{r_i|i=1,2,...N\}$, $\xi$ denotes the set of edges and $A$ the adjacency matrix. We define the connectivity of the graph according to the similarity of the passenger demand patterns among the regions. 
\begin{equation}
A_{i,j}=
\begin{cases}
  1, & \text{if} \  Similarity_{r_i,r_j} > \epsilon \\
  0, & \text{otherwise}
\end{cases}
\end{equation}
where $\epsilon$ is a threshold to which will determine the sparsity of matrix $A$.
\noindent To quantify the similarity in passenger demand between different regions, we use the Pearson Correlation Coefficient. Let $D_{0 \sim t}(r_i)$ represent the historical passenger demand sequence for region $r_i$ from time $0$ to $t$ (in the training data). Then the similarity of region $r_i$ and $r_j$ can be defined as:

\vspace{-0.3cm}
\begin{equation}
    Similarity_{r_i,r_j} = Pearson(D_{0 \sim t}(r_i), D_{0 \sim t}(r_j))
\end{equation}
\vspace{-0.3cm}

\subsection{Long-term and Short-term Encoders}
Most prior works only consider next-step prediction, i.e., predicting passenger demand in the next time step. They optimize these models by reducing the error incurred in the prediction for the next time step in the training stage without considering the subsequent time steps. Hence, these methods are known to deteriorate rapidly for multi-step prediction.
Only a few works have considered the problem of prediction for multiple time steps \cite{ConvLSTM,DCRNN}. These works adopt an encoder-decoder architecture based on RNN or its variants (i.e., ConvLSTM) as the encoder and decoder.
These methods have two disadvantages: 
(1) The chain-structured RNNs employed in the encoder iterate over one input time step at a time. Thus, they require an equivalent number (i.e., $h$) of iterative RNN units when using the historical data over $h$ time steps as the input.
The long calculation distance between target demand and previous demands could cause severe information oblivion. 
(2) In the decoder part, to predict the demand for time step $T$, RNN takes as input the hidden state and the prediction of the previous time step $T-1$. Thus, the errors from the previous time step are carried forward and directly influence the prediction, which results in error accumulation in each future step. 

Different to all these previous works, we introduce an architecture that relies on the long-term and short-term encoder operating simultaneously to achieve multi-step prediction without the use of RNN. The long-term encoder takes the most recent $h$ time steps citywide historical passenger demand sequence \{$\bm{D_{t-h+1}}, \bm{D_{t-h+2}}, ..., \bm{D_t}$\} as input to learn the historical spatial-temporal patterns. These $h$ steps citywide demand are combined and organized into a 3-D matrix, shape $h \times N \times d_{in}$. The long term encoder comprises a number of GCCMs, wherein each GGCM captures spatial correlation among all $N$ regions and temporal correlation among $k$ (the patch size, a hyperparameter) time steps, which we will elaborate on in Section 3.3. Thus, only $\frac{h-1}{k-1}$ iterative steps are needed to capture the temporal correlation over the $h$ historical steps. Compared to a RNN structure, our GGCM-based long-term encoder significantly decreases the iterative steps, which can further decrease the information loss. The output of the long-term encoder is another matrix $Y_h$ shaped $h \times N \times d_{out}$, which is the encoded representation of the input

The short-term encoder is used to integrate already predicted demand for multi-step prediction. It uses a sliding window sized $q$ to capture the recent spatial-temporal correlations. When predicting the passenger demand at time step $T$ ($T \in [t+1, t+\tau]$), it takes the citywide passenger demand of the most recent $q$ time steps, i.e., \{$\bm{D_{T-q}}, \bm{D_{T-q+1}}, ..., \bm{D_{T-1}}$\} as input. Except for the length of time steps ($h$ and $q$ in long-term and short-term, respectively), the operation of the short-term encoder is the same as the long-term encoder. The short-term encoder generates a matrix $Y_q^T$ shaped $q \times N \times d_{out}$ as the near trend representation.
In contrast to a RNN-based decoder, the prediction of the last time step is fed back exclusively to the short-term encoder. Thus, the prediction error will be attenuated by the long-term encoder, which eases the severe error accumulation problem that is inherent in the RNN-based decoder model \cite{STGCN}.

\subsection{Gated Graph Convolutional Module}

\begin{figure}
\includegraphics[height=1.3in, width=3.3in]{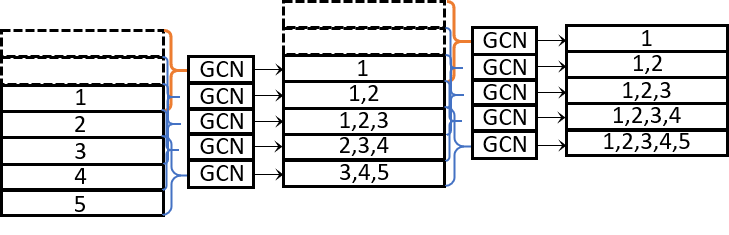}
\vspace{-0.3cm}
\caption{An illustration of extracting spatial-temporal correlations with GCN. }
\label{c2s}
\vspace{-0.2cm}
\end{figure}

The Gated Graph Convolutional Module is the core of both long-term encoder and short-term encoder. Each GGCM consists of several GCN layers, which are \textbf{parallelized along the temporal axis}. To capture the spatial-temporal correlations, each GCN layer operates on a limited historical window ($k$ time steps) of the citywide demand data. It can extract the spatial correlation among all regions within these $k$ time steps. By stacking multiple \textbf{serial} GGCMs, our model forms a hierarchical structure and can capture the spatial-temporal correlations from the entire input ($h$ for long-term encoder and $q$ for short-term encoder, $h > q \geq k$). Figure \ref{c2s} illustrates the exclusive use of GCN for extracting spatial-temporal correlations, where we omit the channel (dimension) axis for simplicity. In literature, there is one similar work \cite{STGCN} to our GGCM module. Their work first employs CNN to capture temporal correlation and then uses GCN to capture spatial correlation. Our approach is significantly simplified compared to theirs, as we can extract the spatial-temporal correlations simultaneously.

\begin{figure}
\includegraphics[height=1.3in, width=3.3in]{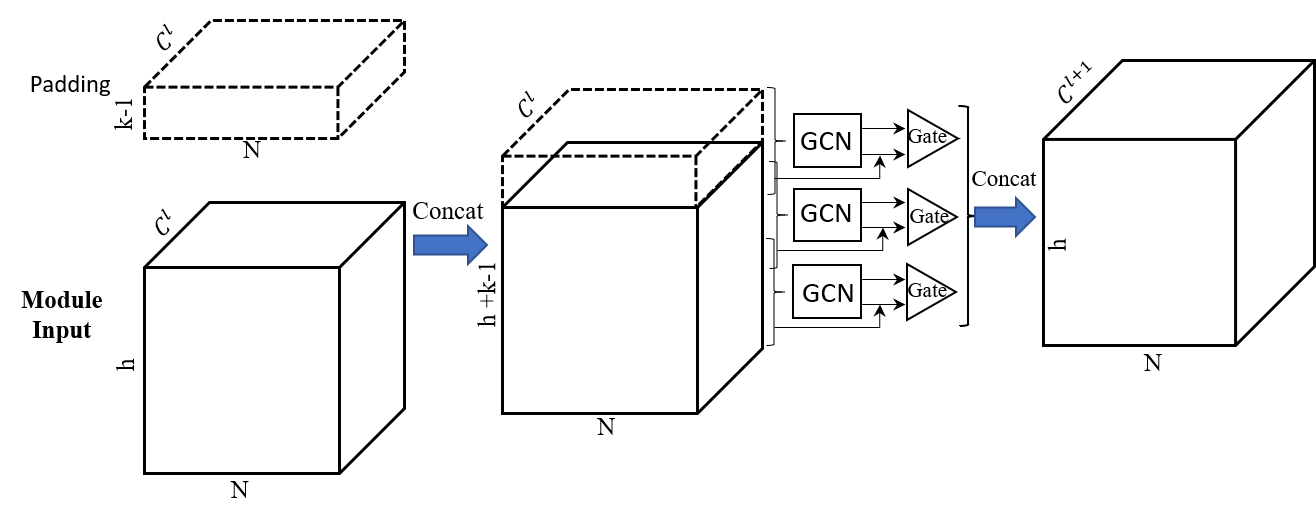}
\vspace{-0.3cm}
\caption{The spatial-termporal Gated Graph Convolutional Module}
\label{GGCM}
\vspace{-0.2cm}
\end{figure}

The detailed design of the GGCM module is shown in Figure \ref{GGCM}. The input of the $l_{th}$ GGCM is a matrix shaped $h \times N \times C^l$ (or $q \times N \times C^l$ for short-term encoder, we only use $h$ in the following for simplicity), where $C^l$ is the input dimension. In the first GGCM module, $C^l$ is $d_{in}$ (as notated in Section 2). The output shape of $l_{th}$ GGCM is $h \times N \times C^{l+1}$. We first concatenate a zero padding matrix shaped $(k-1) \times N \times C^l$ and form the new input $(h+k-1) \times N \times C^l$ to ensure the transformation do not decrease the length of the sequence. Next, each GCN in the GGCM takes $k$ time steps data shaped $k \times N \times C^l$ as input to extract spatial-temporal correlations, which is further reshaped as a 2-D matrix $N \times (k \cdot C^l)$ for GCN calculation. According to \cite{Thomas:Kipf}, the calculation of a GCN layer can be formulated as:

\begin{equation}
    X^{l+1} = (\widetilde{P}^{-\frac{1}{2}} \widetilde{A} \widetilde{P}^{-\frac{1}{2}}) X^l {W}
\end{equation}

where $\widetilde{A} = A + I_n$ ($A$ is the adjecency matrix of the graph defined in Section 3.1, $I_n$ is the identity matrix), $\widetilde{P}_{ii} = \sum_j \widetilde{A}_{ij}$, $X \in \mathbb{R}^{N \times (k \cdot C^l)}$ denotes the reshaped $k$ time steps demand, $W \in \mathbb{R}^{(k \cdot C^l) \times C^{l+1})}$ represents the learned parameters, $X^{l+1} \in \mathbb{R}^{N \times C^{l+1}}$ is the output of GCN.

Furthermore, we adopt the gating mechanism \cite{gate} to model the complex non-linearity in passenger demand forecasting. And Eq. (4) is re-formulated as:

\begin{equation}
    X^{l+1} = ((\widetilde{P}^{-\frac{1}{2}} \widetilde{A} \widetilde{P}^{-\frac{1}{2}}) X^l {W_1} + X^l) \otimes \sigma ((\widetilde{P}^{-\frac{1}{2}} \widetilde{A} \widetilde{P}^{-\frac{1}{2}}) X^l {W_2})
\end{equation}

where $\otimes$ is the element-wise product operation, $\sigma$ denotes the sigmoid function. Thus, the output is a linear transformation $(\widetilde{P}^{-\frac{1}{2}} \widetilde{A} \widetilde{P}^{-\frac{1}{2}}) X^l {W_1} + X^l$ modulated by a non-linear gate $\sigma ((\widetilde{P}^{-\frac{1}{2}} \widetilde{A} \widetilde{P}^{-\frac{1}{2}}) X^l {W_2})$. The non-linear gate controls which part of the linear transformation can pass through the gate and contribute to the prediction. Besides, residual connection \cite{resnet} is utilized to avoid the network degradation as shown in Eq. (5). 

Finally, the $h$ outputs from the gating mechanism are combined along the temporal axis to generate the output of one GGCM module shaped $h \times N \times C^{l+1}$.

\begin{table*}[!htbp]
\caption{Evaluation on next-step prediction over three datasets of different scales (best performance displayed in bold).}
\vspace{-0.2cm}
\label{tabel1}
\centering
\resizebox{\linewidth}{!}{
\begin{tabular}{c|l|ccc|ccc|ccc}
\hline
\multirow{2}{*}{Index} & \multirow{2}{*}{Method} & \multicolumn{3}{c|}{DidiSY} & \multicolumn{3}{c|}{BikeNYC} & \multicolumn{3}{c}{TaxiBJ} \\ \cline{3-11} &
 & RMSE & MAE & MAPE & RMSE & MAE & MAPE & RMSE & MAE & MAPE \\ \hline
1 & HA & 4.112 & 2.646 & 0.426 & 8.541 & 3.695 & 0.437 & 40.439 & 20.696 & 0.268 \\
2 & OLR & 3.713 & 2.528 & 0.379 & 8.502 & 4.652 & 0.391 & 23.921 & 14.937 & 0.276 \\
3 & XGBoost  & 3.612 & 2.394 & 0.402 & 6.914 & 3.423 & 0.367 & 22.927 & 13.687 & 0.212 \\
4 & DeepST  & 3.362 & 2.221 & 0.337 & 6.603 & 2.549 & 0.242 & 18.305 & 11.264 & 0.157 \\
5 & ResST-Net  & 3.449 & 2.331 & 0.318 & 6.159 & 2.432 & 0.228 & 17.649 & 10.599 & 0.141 \\
6 & DMVST-Net  & 3.440 & 2.232 & 0.373 & 4.766 & 2.318 & 0.224 & 18.206 & 11.085  & 0.153  \\
7 & ConvLSTM  & 3.414 & 2.222 & 0.379 & 4.745 & 2.435 & 0.226 & 18.788 & 11.461 & 0.163 \\
8 & FCL-Net  & 3.364 & 2.172 & 0.381 & 4.959 & 2.362 & 0.275 & 18.176 & 10.756 & 0.169 \\
9 & FlowFLexDP  & 3.292 & 2.143 & 0.336 & 6.003 & 2.801 & 0.271 & 19.538 & 11.945 & 0.160 \\
10 & DCRNN  & 3.465 & 2.281 & 0.371 & 5.215 & 2.776 & 0.241 &20.569  &12.517  &0.177  \\ 
11 & STGCN & 3.397 & 2.236 & 0.372 & 4.759 & 2.438 & 0.220 & 19.101  & 11.573  & 0.167  \\ \hline
12 & \textbf{STG2Seq} & \textbf{3.206} & \textbf{2.134} & \textbf{0.306} & \textbf{4.513} & \textbf{2.257} & \textbf{0.210} & \textbf{17.241} & \textbf{10.219} & \textbf{0.138} \\ \hline
\end{tabular}}
\vspace{-0.4cm}
\end{table*}

\subsection{Attention-based Output Module}
As notated in Section 3.2, the long-term spatial-temporal dependency and recent time spatial-temporal dependency are captured and represented as two matrices $Y_h$ and $Y_q^T$ for the target time step $T$. We concatenate them together to form a joint representation $Y_{h+q} \in \mathbb{R}^{(h+q) \times N \times d_{out}}$, which will be decoded by the attention-based output module to obtain the prediction (we omit $T$ here and in the following for simplicity). The three axes of $Y_{h+q}$ are time, space (i.e., region) and channel (i.e., dimension), respectively.

We first introduce the temporal attention mechanism for decoding $Y_{h+q}$. The passenger demand is a typical time series data as previous historical demands have an influence on the future passenger demand. However, the importance of each previous step to target demand is different, and this influence changes with time. We design a temporal attention mechanism to add an importance score for each historical time step to measure the influence. The score is generated by aligning the joint representation $Y_{h+q} = [y_1, y_2, ..., y_{h+q}] (y_i \in \mathbb{R}^{(N \times d_{out})})$ with the target time step's time features $\bm{E_T}$, which can adaptively learn the dynamic temporal influence of each previous time step changing with time. We define the calculation of temporal attention as:

\vspace{-0.3cm}
\begin{equation}
    \bm{\alpha} = softmax(tanh(Y_{h+q} W_3^Y  + E_T W_4^E  + b_1))
\end{equation}

where $W_3^Y \in \mathbb{R}^{(h+q) \times (N \times d_{out}) \times 1}$, $W_4^E \in \mathbb{R}^{d_e \times (h+q)}$ and $b_1 \in \mathbb{R}^{(h+q)}$ are transformation matrices to be learned, $\bm{\alpha} \in \mathbb{R}^{(h+q)}$ is the temporal importance score which is normalized by the softmax function $softmax(\cdot)$. The joint representation $Y_{h+q}$ is then transformed by the importance score $\bm{\alpha}$:

\vspace{-0.3cm}
\begin{equation}
    Y_\alpha = \sum_{i=1}^{h+q} \alpha^i y_i \qquad \qquad \in \mathbb{R}^{N \times d_{out}} 
\end{equation}

Inspired by \cite{scacnn} which showed that the importance of each channel is also different, we further add a channel attention module after the temporal attention module to find the most important frames in $Y_{\alpha} = [\mathscr{Y}_1, \mathscr{Y}_2, ..., \mathscr{Y}_{d_{out}}]$ with $\mathscr{Y}_i \in \mathbb{R}^{N}$. The calculation of channel attention is similar to temporal attention:

\vspace{-0.4cm}
\begin{equation}
    \bm{\beta} = softmax(tanh(Y_\alpha W_5^Y + E_T W_6^E + b_2))  
\end{equation}
\vspace{-0.4cm}
\begin{equation}
    Y_\beta = \sum_{i=1}^{d_{out}} \beta^i \mathscr{Y}_i \qquad \qquad \in \mathbb{R}^{N}
\end{equation}
\vspace{-0.1cm}

where $W_5^Y \in \mathbb{R}^{d_{out} \times N \times 1}$, $W_6^E \in \mathbb{R}^{d_e \times d_{out}}$, $b_2$ are transformation matrices; $\bm{\beta} \in \mathbb{R}^{d_{out}}$ is the importance score for each channel. In the case that target demand dimension is 1, $Y_\beta$ is our predicted passenger demand $\bm{D_T^ \prime}$. When the target demand dimension is 2 (predict both start-demand and end-demand \cite{STDN}), we can simply conduct channel attention for each dimension and concatenate them together to form the predicted passenger demand $\bm{D_T^ \prime}$.

\subsection{Optimization}
The outputs of all the $\tau$ time steps constitute the predicted passenger demand sequence $(\bm{D_{t+1}^ \prime}, \bm{D_{t+2}^ \prime}, ..., \bm{D_{t+\tau}^ \prime)}$. In the training process, our objective is to minimize the error between the predicted and actual passenger demand sequences. We define the loss function as sum of the mean squared error between the predicted and actual passenger demand for $\tau$ time steps, written as:

\vspace{-0.2cm}
\begin{equation}
    \mathcal{L}(W_\theta) = \sum_{t+1}^{t+\tau} \| \bm{D_{T}} - \bm{D_{T}^ \prime} \|_2^2 
\end{equation}

where $W_\theta$ represents all the learnable parameters in the network. These can be obtained via back-propagation and Adam optimizer. Further, we use the teacher forcing strategy in the training stage to achieve high efficiency. Specifically, we always use the true value for short-term encoder instead of the predicted value when training the model.

\section{Experiments}

\subsection{Experimental Setup}

We use three real-world datasets in our comparisons, as detailed below:
\begin{itemize}
    \item DidiSY: This is a self-collected dataset that consists of 1) share car demand data from Didi, the biggest online ride-sharing company in China; 
    2) Time meta, including time of day, day of week,  and holidays; 
    This dataset was collected from Dec 5th, 2016  to Feb 4th, 2017 in Shenyang, a large city in China. Each time step is one hour. We use the data from the last six days for testing while the rest for training. 
    \item BikeNYC \cite{ResST-Net}: The public BikeNYC dataset consists of the bike demand and the time meta. The bike demand covers the shared bike hire and returns data of CityBike in New York from 1 Apr 2014 to 30th Sept 2014. Each time step is one hour. To be consistent with previous works that used this dataset  \cite{DeepST}\cite{ResST-Net}, the last ten days' data are used for testing.
    \item TaxiBJ \cite{ResST-Net}: The public TaxiBJ dataset contains taxi demand in Beijing from 1 Mar 2015 to 30th Jun 2015. Similar to the DidiSY dataset, TaxiBJ contains passenger demand, time meta, and meteorological data. Each time step is 30 minutes. The data of the last ten days is used for testing to keep consistent with previous works. 
\end{itemize}

\begin{table*}[]
\centering
\caption{Evaluation of different variants on Bike\_NYC dataset. }
\vspace{-0.2cm}
\label{variants}
\begin{tabular}{c|c|ccc|ccc|ccc}
\hline
\multirow{2}{*}{Index} & \multirow{2}{*}{Removed Component} & \multicolumn{3}{c|}{RMSE} & \multicolumn{3}{c|}{MAE} & \multicolumn{3}{c}{MAPE} \\ \cline{3-11} 
 &  & step1 & step2 & step3 & step1 & step2 & step3 & step1 & step2 & step3 \\ \hline
1 & Short-term Encoder & 4.509 & \textbackslash{} & \textbackslash{} & 2.304 & \textbackslash{} & \textbackslash{} & 0.211 & \textbackslash{} & \textbackslash{} \\
2 & Temporal Attention & 4.540 & 5.259 & 5.730 & 2.350 & 2.674 & 2.811 & 0.211 & 0.229 & 0.246 \\
3 & Channel Attention & 4.618 & 5.310 & 5.861 & 2.334 & 2.552 & 2.717 & 0.213 & 0.239 & 0.258 \\
4 & Gate Mechanism & 4.576 & 5.222 & 5.658 & 2.314 & 2.507 & 2.664 & 0.213 & 0.232 & 0.246 \\
5 & Teacher Forcing & 4.574 & 5.245 & 5.873 & 2.324 & 2.491 & 2.834 & 0.212 & 0.232 & 0.250 \\ \hline
6 & STG2Seq & 4.513 & 5.209 & 5.497 & 2.257 & 2.452 & 2.555 & 0.210 & 0.228 & 0.240 \\ \hline
\end{tabular}
\vspace{-0.4cm}
\end{table*}

Before feeding the data into the model, categorical features such as hour of day, day of week and holidays are transformed by one-hot encoding. The passenger demand is normalized by Min-Max normalization for training and re-scaled for evaluating the prediction accuracy.
We implemented our model in Python with TensorFlow 1.8. In the experiment, historical passenger demand length $h$ is set to 12, sliding window size $q$ and patch size $k$ are both set to 3.
At each time step, we use three evaluation metrics to evaluate the model: Root Mean Square Error (RMSE), Mean Absolute Error (MAE) and Mean Absolute Percentage Error (MAPE).

\vspace{-0.1cm}
\subsection{Experimental Results}
\subsubsection{Next-step Prediction Comparison}
We first compare our method with three representative traditional baselines: 1) Historical Average (HA); 2) Ordinary Linear Regression (OLR); 3) XGBoost \cite{XGBoost}; and eight discriminative state-of-the-art methods: 4) DeepST \cite{DeepST}; 5) ResST-Net \cite{ResST-Net}; 6) DMVST-Net \cite{DMVST}; 7) ConvLSTM \cite{ConvLSTM}; 8) FCL-Net \cite{Jintao:short-term}; 9) FlowFlexDP \cite{FlowFlexDP}; 10)DCRNN \cite{DCRNN}; 11) STGCN \cite{STGCN}.

Because most state-of-the-art methods can only achieve next-step prediction, we use the first time step result of STG2Seq in this comparison. Table \ref{tabel1} presents the comprehensive comparison results. We observe the following: (1) deep learning methods always outperform non-deep learning methods such as HA, OLSR, and XGBoost, which shows the superiority of deep learning methods in capturing nonlinear spatial-temporal correlations. (2) Our model consistently achieves the best performance and outperforms other methods with a significant margin in all three datasets. More specifically, STG2Seq gains 2.6\%, 4.9\% and 2.3\% relative improvements in RMSE; 0.4\%, 2.6\% and 3.6\% relative improvements in MAE; 3.8\%, 4.5\% and 2.1\% relative improvements in MAPE over the best state-of-the-art methods for the three datasets, respectively. The results indicate that our model can capture more accurate spatial-temporal correlations than the state-of-the-art. 

\vspace{-0.1cm}
\subsubsection{Multi-step Prediction Comparison}
\begin{figure}[!h]
\centering
\subfigure[]{
\includegraphics[width=4cm]{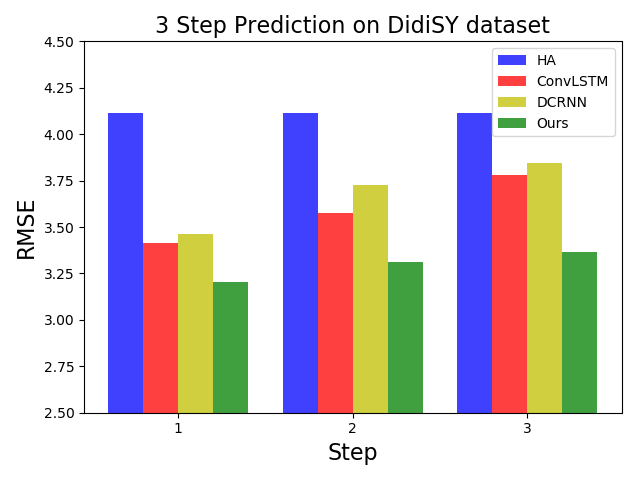}
}
\subfigure[]{
\includegraphics[width=4cm]{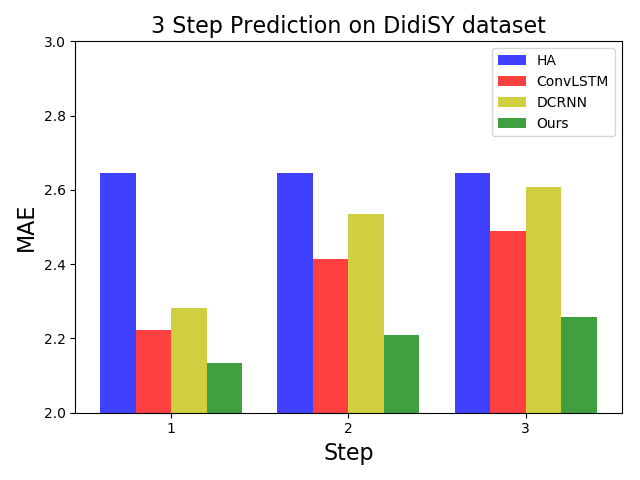}
}
\vspace{-0.3cm}

\subfigure[]{
\includegraphics[width=4cm]{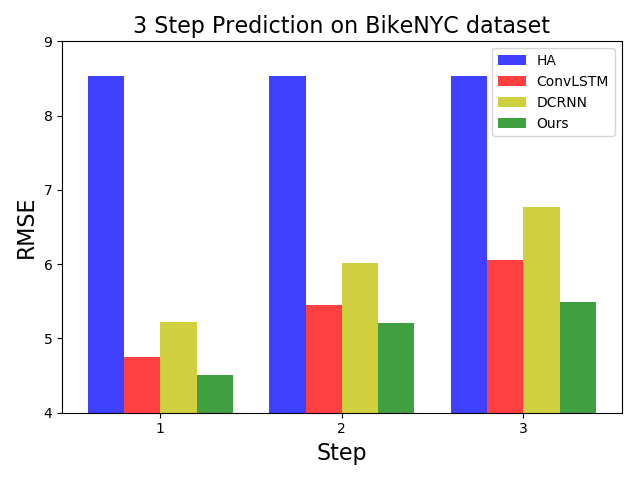}
}
\subfigure[]{
\includegraphics[width=4cm]{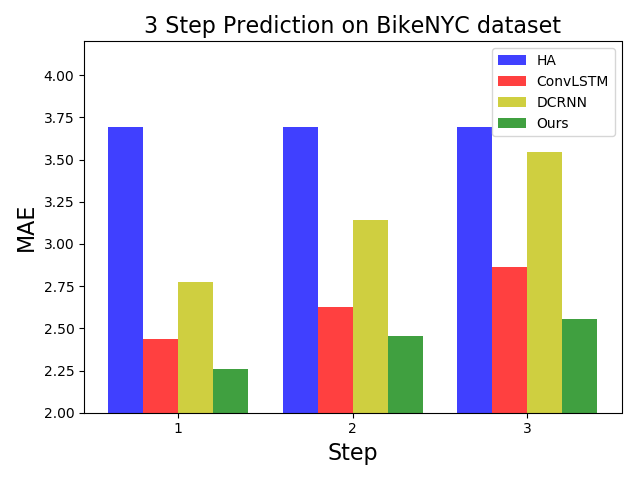}
}
\vspace{-0.3cm}

\subfigure[]{
\includegraphics[width=4cm]{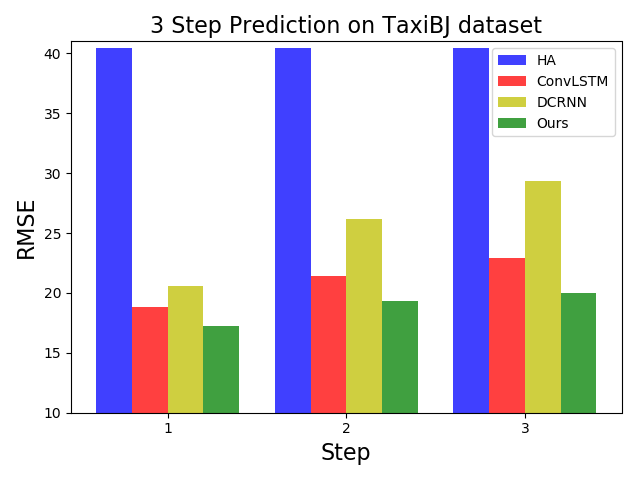}
}
\subfigure[]{
\includegraphics[width=4cm]{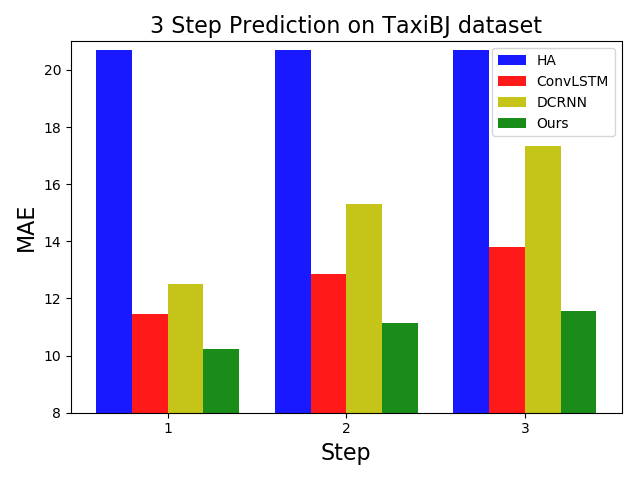}
}
\vspace{-0.3cm}
\caption{Evaluation on multi-step prediction.}
\label{multistep}
\end{figure}

Next, we compare our STG2Seq model with HA, ConvLSTM, and DCRNN, which are also capable of conducting multi-step prediction. Each method predicts the passenger demand in the following $3$ time steps. Figure \ref{multistep} presents the experimental results about RMSE and MAE in three datasets. We can observe that the prediction error for HA is large but consistent for all $3$ time steps. ConvLSTM and DCRNN can achieve good prediction in the first time step. However, they deteriorate fast with time, especially with the DidiSY dataset. STG2Seq can achieve good prediction result for all time steps and deteriorates slower than ConvLSTM and DCRNN, which demonstrates that our method is effective and eases the error accumulation problem that is inherent in RNN-based decoder for multi-step prediction.

\subsubsection{Component Analysis}
To evaluate the effect of different components on our model, we compare five variants obtained by (1) removing the short-term encoder, (2) replacing the temporal attention module by a 2-D CNN layer (for reducing the dimension),  (3) replacing the channel attention module by a 2-D CNN layer, (4) replacing the gate mechanism in GGCM by Relu activation function, and (5) removing the teacher forcing strategy when training the model. The experimental results of each variant are shown in Table \ref{variants}. We gain three observations from this table. First, without the short-term encoder, the model degenerates to a single-step prediction model. Second, the temporal and channel attention modules not only improve the prediction accuracy but also slow down the rate of deterioration in multi-step prediction, which demonstrates the importance of both parts and the effectiveness of our design. Third, the gating mechanism is better at modeling non-linearities when compared with Relu. The results also exemplify the progressive nature of our network design.

\subsubsection{Prediction for Irregular Regions}
All the previous results rely on the assumption that the target regions are partitioned as regular grids. In this final experiment, we investigate the feasibility of our method when the city is partitioned into irregular regions.
The DidiSY dataset contains the precise GPS location of each service requests. This allows us to re-partition the city (Shenyang) into sub-regions based on the road network, which results in irregularly sized partitions. Under this setting, our method is flexible and can be applied to the re-organized dataset without modification. However, most state-of-the-arts introduced in the next-step prediction comparison part cannot be used as CNN-based methods are only suitable to extract Euclidean correlations when the city is partitioned to equally sized grids. To benchmark our method, we include three other better suited comparative methods, namely: Auto-Regressive Integrated Moving Average model (ARIMA), Seasonal Auto-Regressive Integrated Moving model (SARIMA) and Multiple Layer Perceptron (MLP). The results in Table \ref{irregular} show that STG2Seq significantly outperforms the baselines, thus demonstrating its generality.

\begin{table}[]
\centering
\caption{Prediction results on irregular regions}
\vspace{-0.3cm}
\label{irregular}
\begin{tabular}{c|l|cc}
\hline
Index & Method     & RMSE  & MAE   \\ \hline
1     & HA         & 4.231 & 2.714 \\
2     & ARIMA      & 4.001 & 2.681 \\
3     & SARIMA     & 3.937 & 2.619 \\
4     & OLR       & 3.719 & 2.496 \\
5     & MLP        & 3.699 & 2.436 \\
6     & XGBoost    & 3.533 & 2.341 \\
7     & FlowFLexDP & 3.518 & 2.322  \\
8     & DCRNN      & 3.526 & 2.332   \\ 
9     & STGCN      & 3.433 & 2.281        \\  \hline
10     & STG2Seq       & 3.244 & 2.136      \\ \hline
\end{tabular}
\end{table}

\section{Conclusion}
In this paper, we propose a novel deep learning framework for multi-step citywide passenger demand forecasting. We formulate the citywide passenger demand on a graph and employ the hierarchical graph convolution architecture to extract spatial and temporal correlations simultaneously. The long-term encoder and short-term encoder are introduced to achieve multi-step prediction without relying on RNN. Moreover, our model considers the dynamic attribute in temporal correlation by using the attention mechanism. Experimental results on three real-world datasets show that our model outperforms other state-of-the-art methods by a large margin. 

\bibliographystyle{named}
\bibliography{ijcai19}

\begin{thebibliography}{}

\bibitem[\protect\citeauthoryear{Bai \bgroup \em et al.\egroup
  }{2019}]{baipakdd}
Lei Bai, Lina Yao, Salil~S Kanhere, Zheng Yang, Jing Chu, and Xianzhi Wang.
\newblock Passenger demand forecasting with multi-task convolutional recurrent
  neural networks.
\newblock In {\em Pacific-Asia Conference on Knowledge Discovery and Data
  Mining}, pages 29--42. Springer, 2019.

\bibitem[\protect\citeauthoryear{Chen and Guestrin}{2016}]{XGBoost}
Tianqi Chen and Carlos Guestrin.
\newblock Xgboost: A scalable tree boosting system.
\newblock In {\em Proceedings of the 22nd acm sigkdd international conference
  on knowledge discovery and data mining}, pages 785--794. ACM, 2016.

\bibitem[\protect\citeauthoryear{Chen \bgroup \em et al.\egroup
  }{2017}]{scacnn}
Long Chen, Hanwang Zhang, Jun Xiao, Liqiang Nie, Jian Shao, Wei Liu, and
  Tat-Seng Chua.
\newblock Sca-cnn: Spatial and channel-wise attention in convolutional networks
  for image captioning.
\newblock In {\em 2017 IEEE Conference on Computer Vision and Pattern
  Recognition (CVPR)}, pages 6298--6306. IEEE, 2017.

\bibitem[\protect\citeauthoryear{Chu \bgroup \em et al.\egroup
  }{2018}]{FlowFlexDP}
Jing Chu, Kun Qian, Xu~Wang, Lina Yao, Fu~Xiao, Jianbo Li, Xin Miao, and Zheng
  Yang.
\newblock Passenger demand prediction with cellular footprints.
\newblock In {\em 2018 15th Annual IEEE International Conference on Sensing,
  Communication, and Networking (SECON)}, pages 1--9. IEEE, 2018.

\bibitem[\protect\citeauthoryear{Dauphin \bgroup \em et al.\egroup
  }{2017}]{gate}
Yann~N Dauphin, Angela Fan, Michael Auli, and David Grangier.
\newblock Language modeling with gated convolutional networks.
\newblock In {\em International Conference on Machine Learning}, pages
  933--941, 2017.

\bibitem[\protect\citeauthoryear{Gehring \bgroup \em et al.\egroup
  }{2017}]{c2s}
Jonas Gehring, Michael Auli, David Grangier, Denis Yarats, and Yann~N Dauphin.
\newblock Convolutional sequence to sequence learning.
\newblock In {\em International Conference on Machine Learning}, pages
  1243--1252, 2017.

\bibitem[\protect\citeauthoryear{He \bgroup \em et al.\egroup }{2016}]{resnet}
Kaiming He, Xiangyu Zhang, Shaoqing Ren, and Jian Sun.
\newblock Deep residual learning for image recognition.
\newblock In {\em Proceedings of the IEEE conference on computer vision and
  pattern recognition}, pages 770--778, 2016.

\bibitem[\protect\citeauthoryear{Ke \bgroup \em et al.\egroup
  }{2017}]{Jintao:short-term}
Jintao Ke, Hongyu Zheng, Hai Yang, and Xiqun~Michael Chen.
\newblock Short-term forecasting of passenger demand under on-demand ride
  services: A spatio-temporal deep learning approach.
\newblock {\em Transportation Research Part C: Emerging Technologies},
  85:591--608, 2017.

\bibitem[\protect\citeauthoryear{Kipf and Welling}{2017}]{Thomas:Kipf}
Thomas~N Kipf and Max Welling.
\newblock Semi-supervised classification with graph convolutional networks.
\newblock In {\em 5th International Conference on Learning Representations
  (ICLR)}, Apr 2017.

\bibitem[\protect\citeauthoryear{LeCun \bgroup \em et al.\egroup }{2015}]{CNN}
Yann LeCun, Yoshua Bengio, and Geoffrey Hinton.
\newblock Deep learning.
\newblock {\em nature}, 521(7553):436, 2015.

\bibitem[\protect\citeauthoryear{Li \bgroup \em et al.\egroup }{2015}]{second}
Yexin Li, Yu~Zheng, Huichu Zhang, and Lei Chen.
\newblock Traffic prediction in a bike-sharing system.
\newblock In {\em Proceedings of the 23rd SIGSPATIAL International Conference
  on Advances in Geographic Information Systems}, page~33. ACM, 2015.

\bibitem[\protect\citeauthoryear{Li \bgroup \em et al.\egroup }{2018}]{DCRNN}
Yaguang Li, Rose Yu, Cyrus Shahabi, and Yan Liu.
\newblock Diffusion convolutional recurrent neural network: Data-driven traffic
  forecasting.
\newblock In {\em ICLR}, Sep 2018.

\bibitem[\protect\citeauthoryear{Moreira-Matias \bgroup \em et al.\egroup
  }{2013}]{first}
Luis Moreira-Matias, Joao Gama, Michel Ferreira, Joao Mendes-Moreira, and Luis
  Damas.
\newblock Predicting taxi--passenger demand using streaming data.
\newblock {\em IEEE Transactions on Intelligent Transportation Systems},
  14(3):1393--1402, 2013.

\bibitem[\protect\citeauthoryear{Xingjian \bgroup \em et al.\egroup
  }{2015}]{ConvLSTM}
SHI Xingjian, Zhourong Chen, Hao Wang, Dit-Yan Yeung, Wai-Kin Wong, and
  Wang-chun Woo.
\newblock Convolutional lstm network: A machine learning approach for
  precipitation nowcasting.
\newblock In {\em Advances in neural information processing systems}, pages
  802--810, 2015.

\bibitem[\protect\citeauthoryear{Yao \bgroup \em et al.\egroup }{2018}]{DMVST}
Huaxiu Yao, Fei Wu, Jintao Ke, Xianfeng Tang, Yitian Jia, Siyu Lu, Pinghua
  Gong, Jieping Ye, and Zhenhui Li.
\newblock Deep multi-view spatial-temporal network for taxi demand prediction.
\newblock In {\em 2018 AAAI Conference on Artificial Intelligence (AAAI'18)},
  2018.

\bibitem[\protect\citeauthoryear{Yao \bgroup \em et al.\egroup }{2019}]{STDN}
Huaxiu Yao, Xianfeng Tang, Hua Wei, Guanjie Zheng, and Zhenhui Li.
\newblock Revisiting spatial-temporal similarity: A deep learning framework for
  traffic prediction.
\newblock In {\em 2019 AAAI Conference on Artificial Intelligence (AAAI'19)},
  2019.

\bibitem[\protect\citeauthoryear{Yu \bgroup \em et al.\egroup }{2018}]{STGCN}
Bing Yu, Haoteng Yin, and Zhanxing Zhu.
\newblock Spatio-temporal graph convolutional networks: A deep learning
  framework for traffic forecasting.
\newblock In {\em IJCAI}, 2018.

\bibitem[\protect\citeauthoryear{Zhang \bgroup \em et al.\egroup
  }{2016a}]{DeepST}
Junbo Zhang, Yu~Zheng, Dekang Qi, Ruiyuan Li, and Xiuwen Yi.
\newblock Dnn-based prediction model for spatio-temporal data.
\newblock In {\em Proceedings of the 24th ACM SIGSPATIAL International
  Conference on Advances in Geographic Information Systems}, page~92. ACM,
  2016.

\bibitem[\protect\citeauthoryear{Zhang \bgroup \em et al.\egroup
  }{2016b}]{thrid}
Kai Zhang, Zhiyong Feng, Shizhan Chen, Keman Huang, and Guiling Wang.
\newblock A framework for passengers demand prediction and recommendation.
\newblock In {\em Services Computing (SCC), 2016 IEEE International Conference
  on}, pages 340--347. IEEE, 2016.

\bibitem[\protect\citeauthoryear{Zhang \bgroup \em et al.\egroup
  }{2017}]{ResST-Net}
Junbo Zhang, Yu~Zheng, and Dekang Qi.
\newblock Deep spatio-temporal residual networks for citywide crowd flows
  prediction.
\newblock In {\em 2017 AAAI Conference on Artificial Intelligence (AAAI'17)},
  pages 1655--1661, 2017.

\bibitem[\protect\citeauthoryear{Zhou \bgroup \em et al.\egroup
  }{2018}]{AttenConvLSTM}
Xian Zhou, Yanyan Shen, Yanmin Zhu, and Linpeng Huang.
\newblock Predicting multi-step citywide passenger demands using
  attention-based neural networks.
\newblock In {\em Proceedings of the Eleventh ACM International Conference on
  Web Search and Data Mining}, pages 736--744. ACM, 2018.

\end{thebibliography}

\end{document}